%% file: egpaper_final.tex
\documentclass[10pt,twocolumn,letterpaper]{article}
\pdfoutput=1
\usepackage{cvpr}
\usepackage{times}
\usepackage{epsfig}
\usepackage{graphicx}
\usepackage{amsmath}
\usepackage{amssymb}

\usepackage{color}
\usepackage{multirow}
\usepackage{hhline}
\usepackage[breaklinks=true,bookmarks=false]{hyperref}

\cvprfinalcopy % *** Uncomment this line for the final submission

\def\cvprPaperID{****} % *** Enter the CVPR Paper ID here

\begin{document}

%%%%%%%%% TITLE
\title{SizeNet: Weakly Supervised Learning of Visual Size and Fit in Fashion Images}

\author{Nour Karessli \hspace{4mm}  Romain Guigour\`{e}s \hspace{4mm}  Reza Shirvany\\
Zalando SE - Berlin, Germany\\
{\tt\small \{nour.karessli, romain.guigoures, reza.shirvany\}@zalando.de}
}

\maketitle
%\thispagestyle{empty}
%%%%%%%%% ABSTRACT
\begin{abstract}
%Statistical methods mainly rely on \mbox{historical} data of articles purchased and returned. 
Finding clothes that fit is a hot topic in the e-commerce fashion industry. Most approaches addressing this problem are based on statistical methods relying on historical data of articles purchased and returned to the store. Such approaches suffer from the cold start problem for the thousands of articles appearing on the shopping platforms every day, for which no prior purchase history is available. We propose to employ visual data to infer size and fit characteristics of fashion articles. We introduce SizeNet, a weakly-supervised teacher-student training framework that leverages the power of statistical models combined with the rich visual information from article images to learn visual cues for size and fit characteristics, capable of tackling the challenging cold start problem. Detailed experiments are performed on thousands of textile garments, including dresses, trousers, knitwear, tops, etc. from hundreds of different brands. 
\end{abstract}

%%%%%%%%% BODY TEXT
\section{Introduction}
\label{sec:intro}
\input{vSize/sections/intro.tex}
%-------------------------------------------------------------------------
\section{Related Work}
\label{sec:Related}
\input{vSize/sections/related_work.tex}
%------------------------------------------------------------------------
\section{Proposed Approach}
\label{sec:Approach}
\input{vSize/sections/approach.tex}
%------------------------------------------------------------------------
\section{Experimental Results and Discussion}
\label{sec:Results}
\input{vSize/sections/experiments.tex}
%------------------------------------------------------------------------
\section{Conclusion}
\label{sec:conc} 
\input{vSize/sections/conclusion.tex}

{\small
\bibliographystyle{unsrt}
\bibliographystyle{ieee}
\bibliography{egbib}
}

\end{document}

%% file: vSize/sections/intro.tex
Fashion industry has been a major contributor to the economy in many countries. Fashion e-commerce, in particular, has largely evolved over the past few years becoming a major player for delivering competitive and customer-obsessed products and services. 
Recent studies have shown that finding the right size and fit is among the most important factors impacting customers purchase decision making process and their satisfaction from e-commerce fashion platforms~\cite{Pisut2017}. In the context of online shopping, customers need to purchase clothes without trying them on. Thus, the sensory feedback phase about how the article fits via touch and visual cues is naturally delayed, leading to uncertainties in the buying process. As a result, a lot of consumers remain reluctant to engage in the purchase process in particular for new articles and brands they are not familiar with.

To make matters worse, fashion articles including shoes and apparel have important sizing variations primarily due to: 1. a coarse definition of size systems for many categories (e.g small, medium, large for garments) ; 2. different specifications for the same size according to the brand ; 3. different ways of converting a local size system to another, as an example in Europe garment sizes are not standardized and brands don't always use the same conversion logic from one country to another.

A way to circumvent the confusion created by these variations is to provide customers with size conversion tables which map aggregated physical body measurements to the article size system. However, this requires customers to collect measurements of their bodies. Interestingly, even if the customer gets accurate body measurements 
 with the aid of tailor-like tutorials and expert explanations, the size tables themselves almost always suffer from 
 high variance that can go up to one inch in a single size. These differences stem from either different aggregated datasets used for the creation of the size tables (e.g. German vs. UK population) or are due to vanity sizing. The latter happens when a brand  deliberately creates size inconsistencies to satisfy a specific focus group of customers based on age, sportiness, etc. which represent major influences on the body measurements presented in the size tables \cite{ujevic2005, shin2007,faust2014}. 
The combination of the above factors leaves the customers alone to face a highly challenging problem of determining the right size and fit during their purchase journey. 

In recent years, there has been a lot of interest in building recommendation systems in fashion e-commerce with major focus on modeling customer preferences using their past interactions, taste, and affinities~\cite{hu2015,arora2016,bracher2016fashion}. Other work involve image classification~\cite{fashion_classification,DeepFashion}, tagging and discovery of fashion products~\cite{amazon,ebay}, algorithmic outfit generation and style extraction~\cite{outfit}, and visual search that focuses on the problem of mapping studio and street images to e-commerce articles~\cite{tamara,lasserre2018studio2shop}. In this context, only very few research work have been conducted to understand how fashion articles behave from the size and fit perspective~\cite{abdulla2017, guigoures2018hierarchical, sembium2017, Sembium2018}, with the main goal of providing a size advice for customers, mainly by exploiting similarities using article sales and returns data, as detailed in~\autoref{sec:Related}. 
Returns have various reasons such as "don't like the article, article damaged, size problems, etc.". We propose a weakly-supervised~\cite{zhou2017brief} teacher-student approach~\cite{vapnik2013nature, vapnik15learning, wong2016sequence} where we first use article sales and size related returns to statistically model whether an article suffers from sizing issues or conversely has a normal size and fit behaviour. In this context, we don't have access to size and fit expert-labeled data for articles, and thus, only rely on weakly-annotated data from the returns process. We then make use of a teacher-student approach with curriculum learning~\cite{bengio2009curriculum} where the statistical model acts as the teacher and a CNN-based model, called SizeNet, acts as the student that aims to learn size issue indicators from the fashion images without direct access to the privileged sales and returns data.

The contributions of our work are three-fold: 1. We demonstrate, for the first time to our best knowledge, the rich value of fashion images in inferring size characteristics of fashion apparel; 2. At the same time our approach is novel in using the image data to effectively tackle the cold start problem that is known to be a very challenging topic in the literature; 3. We propose a teacher statistical model that uses crowd's subjective and inaccurate feedback (highly influenced by personal perception of article size) to generate large scale confidence-weighted weak annotations. This enables us to control the extent to which the weak annotations influence the quality of the final model, and we demonstrate that not applying this approach, i.e. treating weak labels uniformly, highly degrades the quality of the learned model.

The outline of the paper is as follows. In~\autoref{sec:Related} we present related work. In~\autoref{sec:Approach} we present the proposed approach; ~\autoref{sec:teacher} presents the teacher-student framework,~\autoref{subsec:srr} presents the statistical model predicting size issues taking into account the article's category, sales period, number of sales, and number of returns due to size problems. In~\autoref{subsec:NN} we introduce the architecture of the SizeNet along with the curriculum learning approach using the statistical class labels and their confidence scores to train SizeNet on fashion images. In~\autoref{sec:Results} we present two baselines, experimental results, and discussion to assess the quality of the SizeNet results  over different categories of garments including dresses, trousers, knitwear, and tops/blouses. Furthermore, we analyze different cases going from warm to cold start. Finally in~\autoref{sec:conc}, we draw conclusions and discuss future work directions.

%% file: vSize/sections/related_work.tex
The topic of understanding article size issues, and more generally predicting how e-commerce fashion articles may fit customers is challenging. Recent work has been done for supporting customers by providing size recommendations in \cite{abdulla2017} and \cite{guigoures2018hierarchical}. Both approaches propose a personalized size advice to the customer using different formulations. The first one uses a skip gram based word2vec model~\cite{word2vec} trained on the purchase history data to learn a latent representation for articles and customers in a common size and fit
space. Customer vector representation is obtained by aggregating over purchased articles, and a gradient boosted classifier predicts the fit of an article to a specific customer. The second publication proposes a hierarchical Bayesian approach to model what size a customer is willing to buy along with the resulting return status (article is kept, returned because it's too big, or returned because it's too small). 

Following a different approach, the authors of \cite{sembium2017} propose a solution, using the purchase history for each customer, to determine if an article of a certain size would be fit or would suffer a size issue (defined as large, or small). This is achieved by iteratively deducing the true sizes for customers and products, fitting a linear function based on the difference in sizes, and performing ordinal regression on the output of the function to get the loss. Extra features are simply included using addition to the linear function. To handle multiple personas using a single account, hierarchical clustering is performed on each customer account before doing the above. An extension of that work proposes a Bayesian approach on a similar model \cite{Sembium2018}. Instead of learning the parameters in an iterative process, the updates are done with mean-field variational inference with Polya-Gamma augmentation. This method therefore naturally benefits from the nice advantages of Bayesian modeling - the uncertainty outputs, and the use of priors. It however does not tackle the cold-start problem - where zero or very few article sales and returns are available. 

In the fashion e-commerce context, everyday thousands of articles are introduced to the catalog. The life cycle of most articles in fashion e-commerce is usually short - after few weeks, the article is out of stock and removed from the assortment. The "hassle-free" return policy of e-commerce platforms allows customers to return items with no additional cost, whenever they desire up to multiple-weeks from the purchase date. When customers return an item, they can disclose a return reason, for example "did not like the item", "item did not fit" or "item was faulty". In this work, we are interested in estimating whether an item has sizing issues, therefore, we make use of the weakly-annotated size related return data where a customer mentions that an article is not fitting. It is important to note that article returns can reach the warehouses only after multiple days (if not weeks) from the date of the article activation resulting in a cold start period. 

Indeed, all the aforementioned publications state that data sparsity and cold start problem are the two major challenges in such approaches. They propose to tackle those challenges with limited success by exploiting article meta data or taxonomies in the proposed models. In this paper, we leverage the potential of learning visual cues from fashion images to better understand the complex problem of article size and fit issues, and at the same time, provide insights on the value of the image-based approach in tackling the cold start problem. A major advantage of images over meta data or taxonomies lies in the richness of the imagery data, in addition to a lower subjectivity of the information when compared to the large list of ambiguous fashion taxonomies- for example, a slim jeans from Levi's does not follow the same physical and visual characteristics as a slim jeans from Cheap Monday, as both brands target different customer segments. 

%% file: vSize/sections/approach.tex
In this section, we explain the details of the proposed approach to infer article sizing problems from images. We first start by introducing our weakly-supervised  teacher-student framework. Then we introduce our statistical model, and finally we discuss the SizeNet - our CNN model capable of predicting size issues from fashion images thanks to the insights from the statistical model.  

\begin{figure*}[t]
   \centering
    \includegraphics[width=0.85\linewidth]{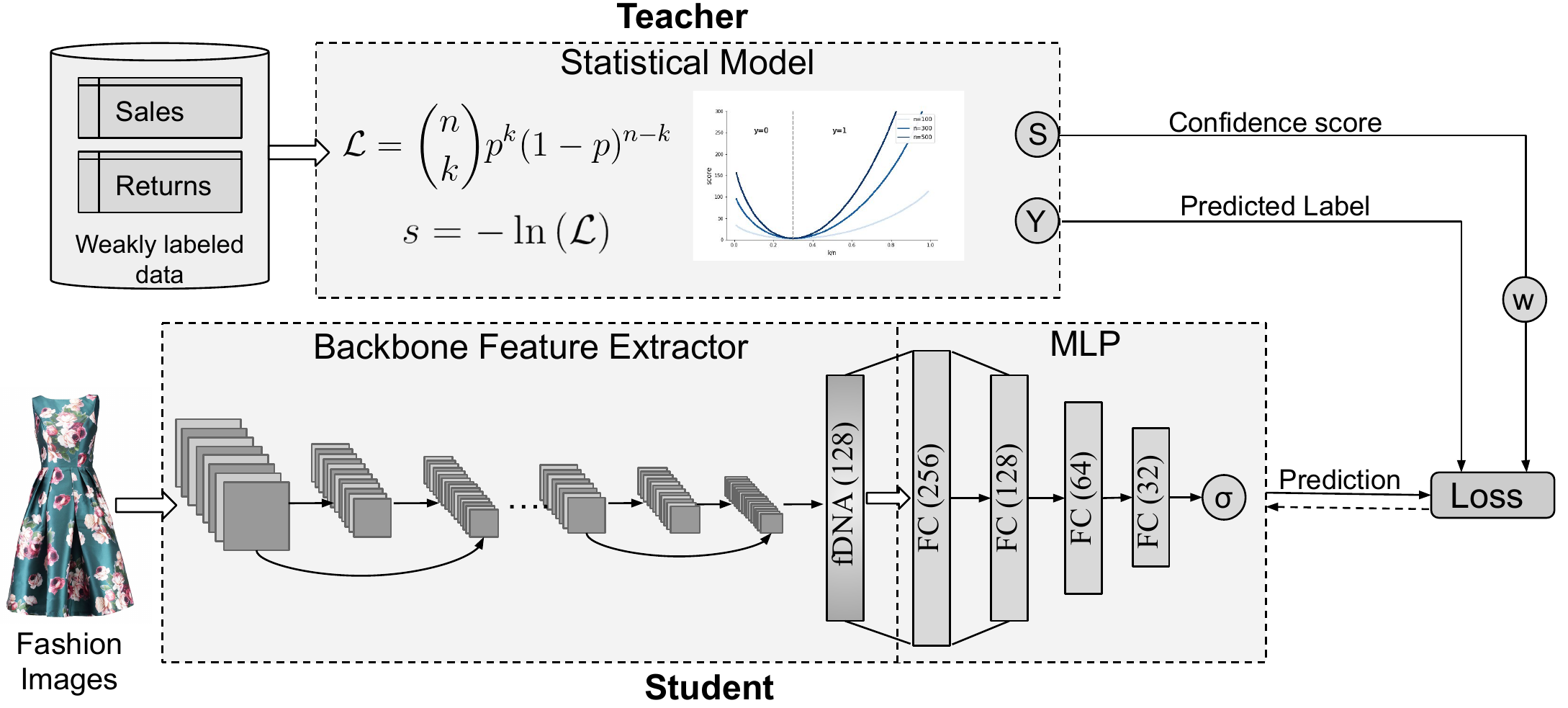}
   % \vspace{2mm}
\caption{{Architecture of the proposed teacher-student approach. On the top, the statistical model acts as the teacher with direct access to the privileged sales and returns data. On the bottom, SizeNet is shown as the student, composed of a CNN backbone feature extractor followed by a multi-layer perceptron.}}
\label{fig:network}
\end{figure*}
\subsection{Teacher-Student Learning}
\label{sec:teacher}
The concept of training machine learning models following a teacher-student approach is a well-known concept where its mention in the community dates back to 90s. In recent years, however, there has been an extensive interest in further developing the teacher-student and related learning frameworks such as the curriculum learning approaches~\cite{vapnik15learning, bengio2009curriculum, wong2016sequence}. Interestingly, to motivate the teacher-student training approach, \cite{vapnik15learning} illustrates a cold start problem using the example of an outcome of a surgery three weeks after it has occurred. The classifier is trained on historical data where the historical data contains privileged information about the procedure and its complications during the three weeks following the surgery. The model trained using the privileged data is then considered as the teacher. A second model is trained on the same samples but without using the privileged information. Therefore, this second model - the student - tries to learn from the insights given by the teacher to replicate the outcome of the teacher without directly having access to the privileged data. \cite{vapnik15learning} uses the teacher-student approach along with a support vector machine (SVM) \cite{vapnik2013nature}. In the non-separable case - i.e when there exists no hyperplane separating classes - SVM needs to relax the stiff condition of linear separability and allow misclassified observations.   As shown in \cite{lopez2015unifying}, the teacher also helps refining the decision boundary and can be assimilated to a Bayesian prior.

Following a similar concept, curriculum learning \cite{bengio2009curriculum} suggests to train classifiers first on easier (or more confident) samples and gradually extend to more complex (or less confident) samples. It is shown that this strategy leads to better models with increased generalization capacities. 
Most approaches using a teacher-student learning strategy \cite{vapnik15learning, bengio2009curriculum, wong2016sequence} derive the importance of the samples from the teacher classifier. In this paper, we build a statistical model that has privileged information on the article sales and returns data, as the teacher, and train the SizeNet model, as the student, on fashion images using a confidence score to weigh the samples from the teacher. In other words, the approach is transferring knowledge from privileged information space to the decision space. Though the teacher model in our case does not use the article images as input, it leverages the privileged historical data of sold articles (privileged information space), and the student uses this knowledge to learn from images in the decision space. The confidence-weighted annotations generated by the teacher enables us to control the extent to which these weak annotations (built from the crowd's subjective and inaccurate feedback) influence the quality of the final model, and thus, delivering better learned model.

\input{vSize/sections/size_related_returns.tex}

\subsection{SizeNet: Learning Visual Size Cues}
\label{subsec:NN}
In this section, we propose the SizeNet architecture to investigate the article size and fit characteristics in a weakly-supervised teacher-student manner using fashion images. We make use of the labels and their confidence scores acquired from the statistical model described in the previous section, to teach the image-based SizeNet model size issue classification. In particular, we adopt a curriculum learning  approach that gradually makes use of feeding the articles with high size issue confidence scores for learning confident visual representations for sizing issues in the images following by less confident samples to improve generalization. \autoref{fig:network} illustrates the architecture of our approach including the statistical model, and the proposed SizeNet composed of a CNN backbone feature extractor followed by a multi-layer perceptron. %
% \vspace{-4mm}
\paragraph{Backbone Feature Extractor:} We use the Fashion DNA (fDNA)~\cite{bracher2016fashion} network as a backbone features extractor for SizeNet. The adopted fDNA architecture is similar to a ResNet~\cite{resnet} architecture. The network is pre-trained on 1.33 million fashion articles (sold from 2011 to 2016) with the aim of predicting limited fashion article metadata such as categorical attributes, gender, age group, commodity group, and main article color. Using the fDNA backbone we are able to extract for each image a bottleneck feature vector of dimension 128. 
% \vspace{-4mm}
\paragraph{Multi-Layer Perceptron:} On top of the backbone network, we attach a multi-layer perceptron (MLP) that consists of four fully-connected layers. We opt for a bottleneck MLP approach~\cite{vu2012multilingual} going up from the 128 extracted feature vector, to 256 units and down again to 128. Therefore, the numbers of units of the four fully connected layers are respectively 256, 128, 64 and 32. Each of these layers is followed by a non-linear ReLU activation function. To avoid over-fitting on the training data, we use standard dropout layers for each fully connected layer. The output layer has a sigmoid activation with a unit indicating the sizing issue. We use a binary cross-entropy loss function and optimize the network weights through stochastic gradient descent (SGD). 

We adopt a curriculum that gradually trains the SizeNet starting from more confident samples, coming from the statistical model, down to less confident samples. To prepare the loss function for samples where the label confidence from the statistical model is low, we propose to use a weighting function in the loss. Let's define a sample weight $w_i$ using logarithmic transformation of the sample confidence score $s_i$ as follow:  
\begin{equation}
        w_i = \ln{(1+s_i)}
    \label{eq:weight}
\end{equation} 
The logarithmic transformation allows us to reduce the skewness in the confidence score distribution and provides numerically well-behaving weights compared to the scores. 

Once the network has been fully trained, we evaluate the performance using unseen test data in the next section. We analyze on cases that extend from a. articles where the statistical model provides quality predictions of sizing issue, to b. articles where the statistical model fails to provide quality predictions. The aim of this approach is twofold: first to see to what extent SizeNet is capable of producing quality results comparable to the statistical model using purely images, and second to see to what extent SizeNet can generalize its predictions thanks to the learned visual representations, to those unknown, cold start, and low confidence articles.

%% file: vSize/sections/size_related_returns.tex
\subsection{Statistical Modeling}
\label{subsec:srr}

In this work, we opt for a simplifying approach and formulate the sizing problem as a binary classification problem. Thus, we arrange articles based on their sizing behavior into two categories. Class 1 groups articles that are annotated as having a size issue, e.g. too small, shoulder too tight, sleeves too long, etc. Class 0 groups  other articles with no size issue.
To allocate articles to the appropriate class, we need to consider two factors:
\begin{itemize}
    \item \textit{The category:} Article categories are diverse; some example are shoes, t-shirts, dresses, trousers, etc. Generally, for each category we expect a different return rate and sizing issues. As an example, high heels have a higher size related return rate than sneaker shoes, since customers are more demanding in terms of size and fit with the former than the latter. Therefore, we should consider for each category the amount of size related returns in the category compared to that of its average.

    \item \textit{The sales period:} The usual life cycle of an article starts with its activation on the e-commerce platform, after which customers start purchasing the article and potentially return the article if it does not meet their expectations. This process naturally results in a time dependency in the purchase and return data. Therefore, for each category, we should consider the amount of the size related returns of an article compared to the amount of the returns in its category over the same time period.
\end{itemize}

Therefore considering the above points, if an article has higher size related returns than the average of its category over the same period of time, then the article is considered to demonstrate a sizing problem (labeled as class~1); otherwise, it is considered to have a normal sizing characteristics and thus belongs to the no-sizing-issue class (labeled as class~0).   

For each article and category our confidence in labeling the article as a size issue or not greatly depends on how large the number of sales and returns are. Therefore, we propose to use a binomial likelihood $\mathcal{L}$ to assess the confidence in the class assertion. Let's denote $p$ the expected size related return rate of the item, i.e the size related return rate of its category, $k$ the number of size related returns of the item, and $n$ the number of purchases. We can define the binomial likelihood as following:

\begin{equation}
    \mathcal{L} = \binom{n}{k} p^k (1-p)^{n-k}
\end{equation}

We note that, the value of the likelihood is maximized when the ratio of $k$ over $n$ is equal to $p$. In other words, $k$ is the expected number of size related returns sampled from the distribution $p$, when drawing $n$ times. The more observations are sampled, the more the estimator is confident. That way, for a large value of $n$, if the ratio $k$ over $n$ diverges from $p$, the likelihood is low. Conversely, if only few observations have been sampled, the estimator is really uncertain and tends to distribute the density over all possible values of $k$. Let's define a score $s$ based on the negative logarithm of the binomial estimator:
\begin{equation}
    s = -\ln{(\mathcal{L})}
    \label{eq:est_score}
\end{equation}

\begin{figure}[!ht]
   \centering
    \includegraphics[width=\linewidth]{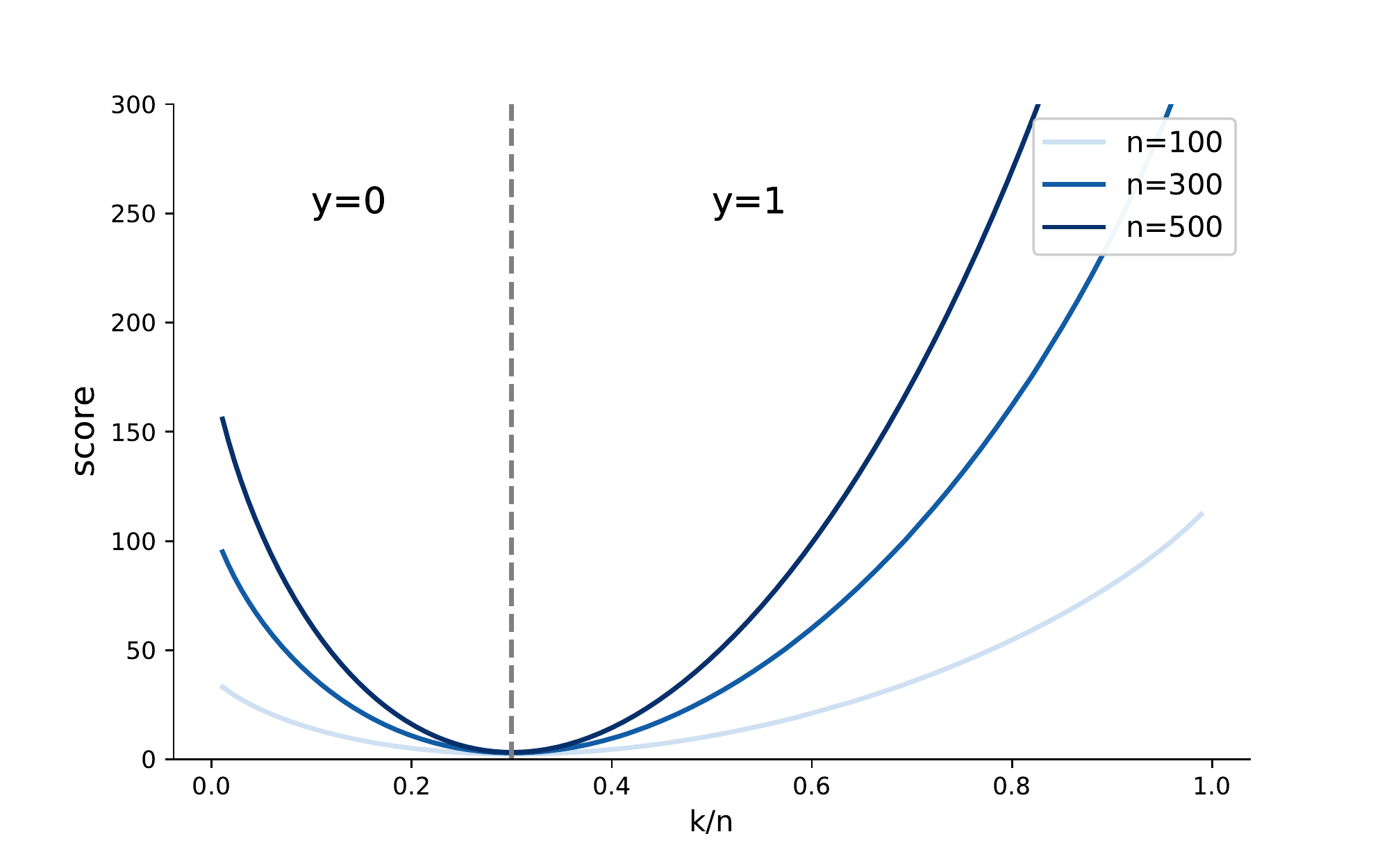}
\caption{Y-axis is the value of the score $s$. X-axis is the ratio of the number of size related returns $k$ over the number of sales $n$. Curves are plotted for different amounts of sales $n$. In this example the expected size return rate $p$ is arbitrarily set at $0.3$ for illustration (vertical dashed gray line). }
   \label{fig:binomial}
\end{figure}

In that way, the score $s$ is very high when $k$ is unlikely to have been sampled from $p$, meaning that the size related return rate is either very high (sizing problem, class $y=1$) or very low (no size issue, class $y=0$). ~\autoref{fig:binomial} shows the behaviour of $s$ with respect to $n$ and $k$. In this Figure, as an example, let's assume that the expected size return rate $p$ for a defined category and a fixed sales period is $0.3$ (vertical dashed gray line). Therefore, articles in this category and in the same sales period, for which the size related returns is larger than $0.3$ (right side of the line) are considered to demonstrate a sizing problem (labeled as $y = 1$). On the other hand, for the same ratio of $k$ over $n$, we see how an increase in number of purchases $n$ (different $U$ shape blue curves) results in an increase in the score $s$, and thus,  demonstrating a better confidence in the class assertion.

To get a better understanding of the score function $s$, we can look at the asymptotic interpretation of the \autoref{eq:est_score}. By applying the Stirling approximation, we can easily derive the following property:
\begin{equation}
    s \rightarrow n KL (\bar{P} || P) \mbox{ when } n \rightarrow \infty
    \label{eq:weight}
\end{equation}
where $P=\{p, 1-p\}$ and $\bar{P}=\{\frac{k}{n}, 1-\frac{k}{n}\}$, and $KL$ denotes the Kullback-Leibler divergence~\cite{kullback1951information}. This property provides a better understanding of the behaviour of the confidence score: if $\bar{P}$ is very different from $P$, i.e if the size related return rate is either way lower or way higher than the one of its category, then the Kullback-Leibler divergence is high and $s$ is high too. However, if the number $n$ of purchases of the article is low, the score is penalized. 

The negative log likelihood - as well as the Kullback-Leibler divergence - are defined on $\mathbb{R}^+$, and consequently can in theory tend to the infinity. In practice, we can however define upper bounds for the score $s$. Upper bounds are reached when $p$ is very different from the ratio $k$ over $n$, i.e in the following two cases:
\begin{itemize}
    \item when 
    $p \rightarrow 0^+$ and $k=n$, then $s = -n\ln({p})$
    \item when $p \rightarrow 1^-$ and $k=0$, then $s = (k-n)\ln{(1-p)}$
\end{itemize}

Note that the cases $p=0$ and $p=1$; that is when the size related return rate of the category is zero, or in contrast, when all items are returned, define very interesting edge cases. The first usually happens in the few weeks that follow the activation of the articles on the e-commerce platform, where no returns are recorded yet.  Therefore, in this case $p=0$ implies $k=0$; since as soon as we record a return for an article, we also record a return for its category. As a consequence, for this case the binomial likelihood is equal to $1$, meaning that the confidence score is zero. Therefore, the statistical model is not capable of providing any size issue prediction. It is important to note that this case actually corresponds to the challenging cold start problem in e-commerce fashion for which we propose a solution in this paper thanks to our SizeNet approach. The latter case where $p=1$ implies $k=n$, and the confidence score is also zero. However, this scenario, in which all articles are returned due to size issues, is practically non-existent in the e-commerce context. 

Now that we have established our statistical model as the teacher, capable of providing sizing class labels with a confidence score, we discuss the student for learning of visual cues for size issues following a curriculum learning framework, keeping in mind the generalization to the cold start articles.

%% file: vSize/sections/experiments.tex
\label{sec:experiments}
In this section, we conduct multiple experiments to evaluate and understand the performance of the proposed SizeNet model over multiple garment categories from around $500$ different brands. 

\subsection{Dataset} For our experiments, we use an in-house rich dataset of women textile garments including $12$ categories such as dresses, blouses, jeans, skirts, jackets, etc. collected from around $500$ different brands. Observations are defined at the stock keeping unit (SKU) level. This means that two pieces of garments belonging to the same model, but with different colors, are considered as two distinct observations. We justify this choice by two main reasons derived from expert knowledge: 1. manufacturers use different fabrics depending on the dying technique, 2. customers don't perceive size and fit the same way depending on the color of clothes. Those two points lead to very different size related return reason distributions for the same article model but with different colors.

{
\renewcommand{\arraystretch}{1.2}
\begin{table}[!ht]
 \begin{center}
    \begin{tabular}{lcc}
    \textbf{Class} & \textbf{\#Articles} & \textbf{\# Images}  \\     \hline
    size issue & 68,892 & 69,064 \\ 
    no size issue & 58,152 & 58,321 \\ \hline 
    total & 127,044 & 127,385 \\ \hline
  \end{tabular}
 \end{center}
 \caption{Overall statistics of used women textile dataset, showing the number of SKUs and the number of related images in each class according to the statistical model labels (\autoref{subsec:srr}).}
    \label{tab:datasets}
\end{table}
}

The dataset in-hand was composed of a relatively balanced size-issue/no-size-issue subset of the articles as reported  ~\autoref{tab:datasets}. The class labels and confidences are derived from the statistical model described in~\autoref{subsec:srr}. Articles activated in the last 6 months were excluded in this dataset to ensure the quality of the return data. We opt to use packshot images with white background and without a human model. 
% Some articles have multiple images of different views e.g. front, side and back views. 
We do not perform any task-specific image pre-processing, the input images are simply re-sized to $177 \times 256$. 
The data set was split by maintaining a ratio of 60/20/20 for training, validation, and test sets respectively. We cross-validate hyper-parameters of the network, such as start learning rate, batch size, number of epochs, and stopping criteria using the validation set. 
\subsection{Evaluation}
\begin{figure}%[!ht]
  \centering
    \includegraphics[width=\linewidth]{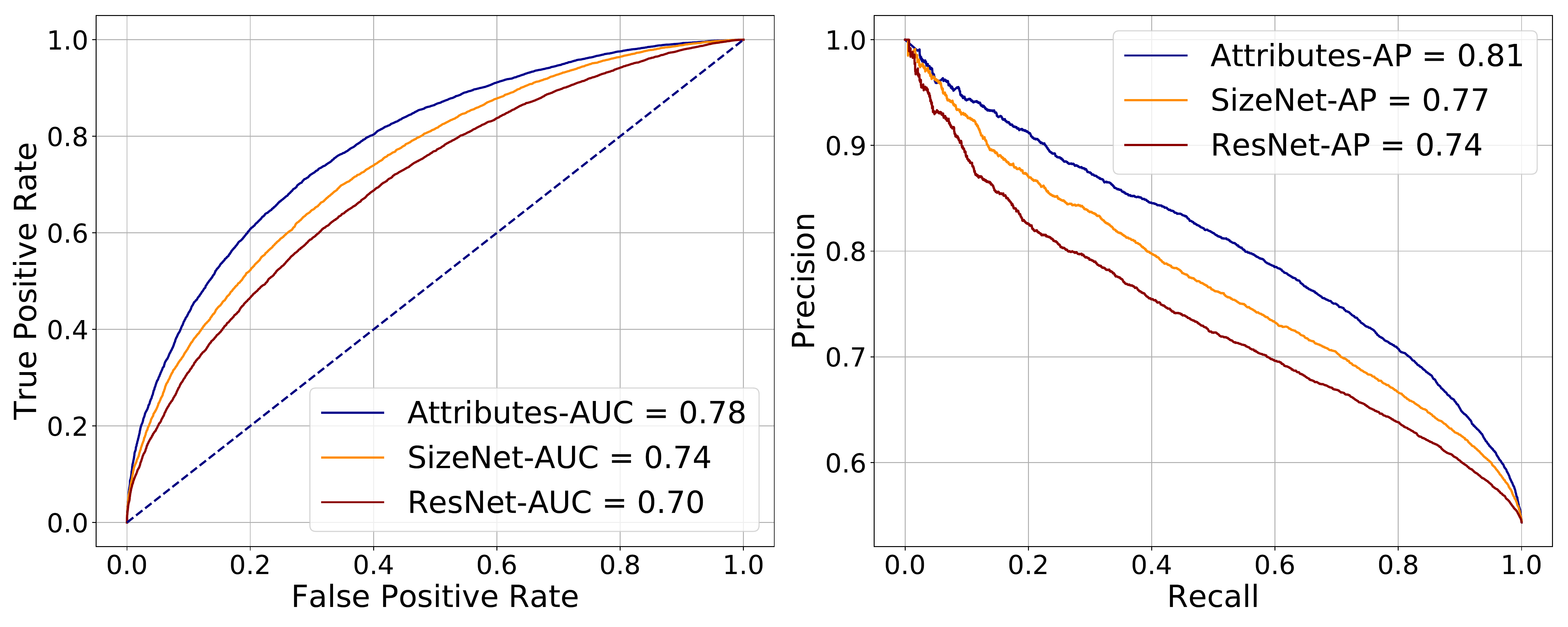}
\caption{Evaluation of size issue prediction for the overall dataset (12 categories) comparing SizeNet to two baselines. Left: Receiver Operating Characteristic (ROC) curves with area under curve (AUC); Right: Precision-Recall curves with average precision (AP).}
\label{fig:baseline}
\end{figure}

\begin{figure}
   \centering
   \includegraphics[trim={25.5cm 21.2cm 0 0},clip,width=0.49\linewidth]{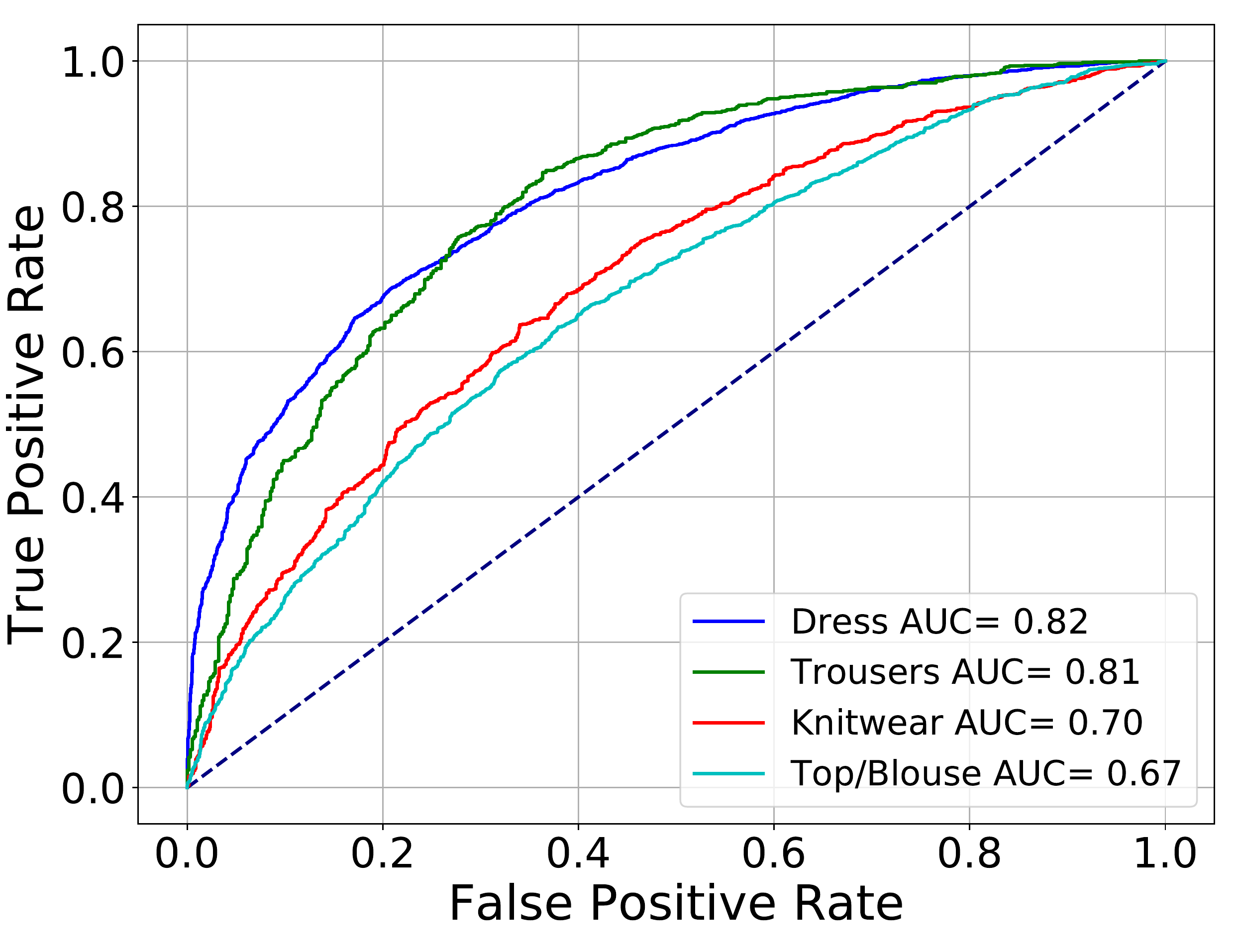}
    \includegraphics[trim={25.5cm 0 0 21cm},clip,width=0.49\linewidth]{vSize/figs/roc-pr-all.pdf}
\caption{Evaluation of size issue prediction for the four major categories. Left: Receiver Operating Characteristic (ROC) curves with area under curve (AUC); Right: Precision-Recall curves with average precision (AP).}
\label{fig:curves}
\end{figure}

In order to assess the performances of our model, we first study the classification metrics including the receiver operating characteristic (ROC) and precision-recall curves.
\paragraph{Baselines:} 
We introduce two baselines: first baseline is a model denoted as \texttt{Attributes} that instead of article image uses sparse k-hot encoding vector of binary fashion attributes (e.g. fabric material, fit type, length, etc.) of size 13,422. These attributes are created following a laborious and costly process by human expert annotators. As a second baseline, we use a standard \texttt{ResNet} pretrained on ImageNet as the backbone CNN instead of fDNA. We report the results for the overall size issue predictions ($12$ categories combined). \autoref{fig:baseline} demonstrates that \texttt{SizeNet} outperforms \texttt{ResNet} baseline, and achieves promising results compared to that of \texttt{Attributes} model which requires tremendous annotation effort. This benchmark establishes the value of the \texttt{SizeNet} purely using image data.
~\autoref{fig:curves} presents SizeNet performance per category curves for the four major garment categories: dresses, trousers, knitwear, and tops/blouses, where for each category more than $2000$ articles are present in the test set. From these curves we can observe good results for SizeNet predictions; in particular prediction of size issues in dress and trouser categories outperforms other categories.

\begin{figure}[ht!]
   \centering
    \includegraphics[width=0.48\linewidth]{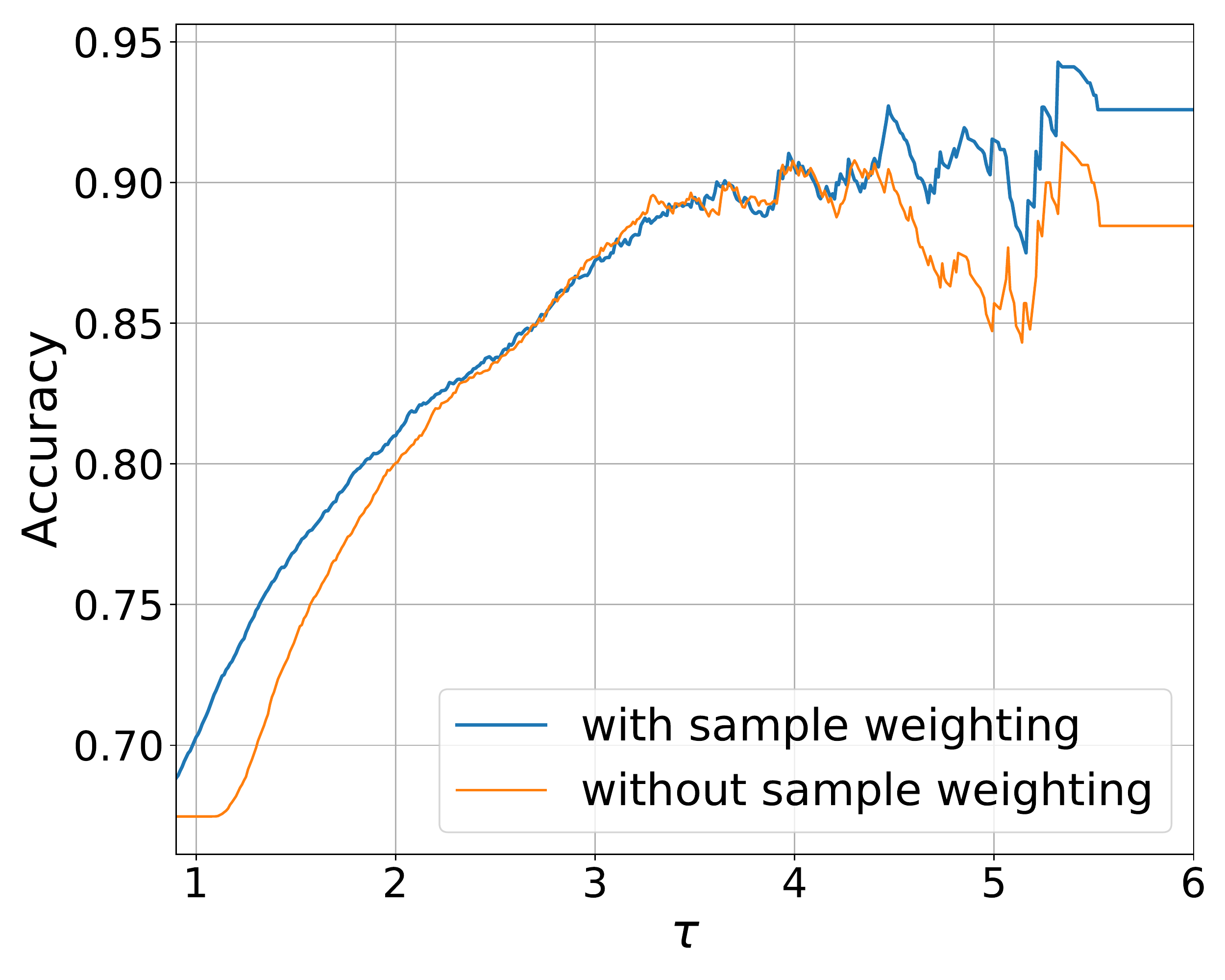}
    \includegraphics[width=0.48\linewidth]{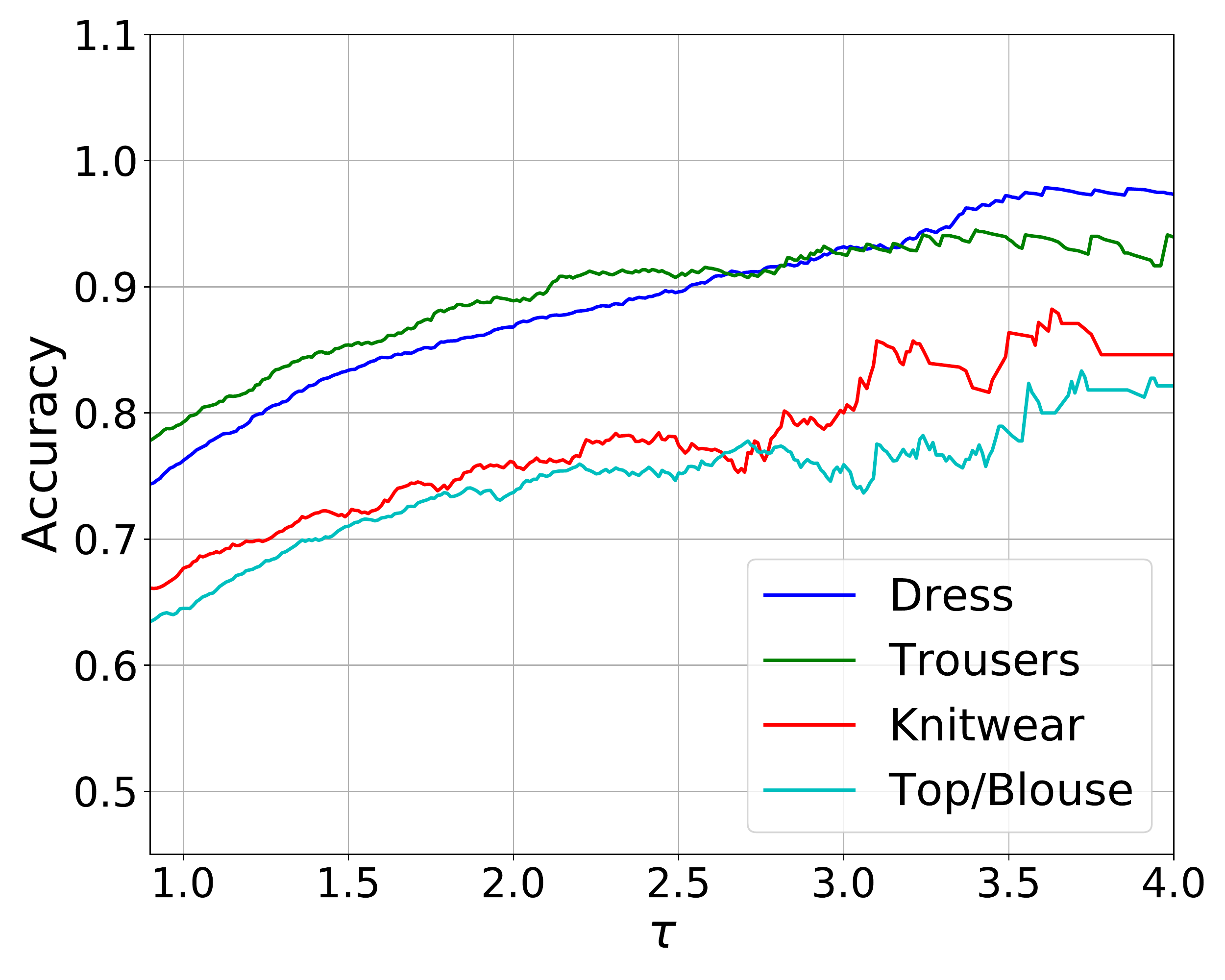} 
\caption{Accuracy of SizeNet model for different thresholds $\tau$ on the test set. Lower values of $\tau$ correspond to including cold-start articles, where an increase in $\tau$ corresponds to only considering articles with larger sales and returns (Left) Overall accuracy with and without using the sample weights $w_i$ in the training phase. (Right) Per category accuracy for major categories using sample weighting in the training phase. }
\label{fig:tau-acc}
\end{figure}

Let's investigate how the teacher and the student interact with each other. As mentioned in~\autoref{sec:Approach}, the neural network (the student) learns the image based size issue predictions from the output of a statistical model (the teacher) that has access to privileged sales and returns data. Samples are weighted to favor regions in the parameter space where the certainty of the teacher is maximal. As a result, we expect to observe good predictions from the student for samples where the teacher is confident. To verify this hypothesis, we plot in~\autoref{fig:tau-acc} the accuracy of the SizeNet model with respect to different values of a threshold $\tau$ applied on the weights $w_i$ which correspond to a monotonous transformation of class confidences from the statistical model. ~\autoref{fig:curves} (left) shows the overall accuracy ($12$ categories) on the test set, obtained both with and without sample weighting during the training phase.~\autoref{fig:curves} (right) shows per category accuracy for four major categories using sample weighting in the training phase. Low values of $\tau$ correspond to all articles particularly including those that suffer from the cold start problem. With higher values of $\tau$, only those articles which are not suffering from the cold start problem are considered (higher confidence in the class).  As expected, the curve shows a high correlation between the SizeNet model performances and the confidence level based on the binomial estimator. 

With regards to the added value of the weighting during the training phase, from~\autoref{fig:tau-acc} (left) we observe that performances of both cases follow the same trend, though for lower values of $\tau$, using the weights in the training phase improves the performances on the test set. For high values of $\tau$, results do not provide much insights since the variance is too high (caused by too few samples). The algorithm exploiting weights is relaxed around the decision boundary in agreement with the study from \cite{vapnik15learning}, leading the model to provide a better generalization capacity.
\begin{figure}[!ht]
  \centering
    \includegraphics[width=\linewidth]{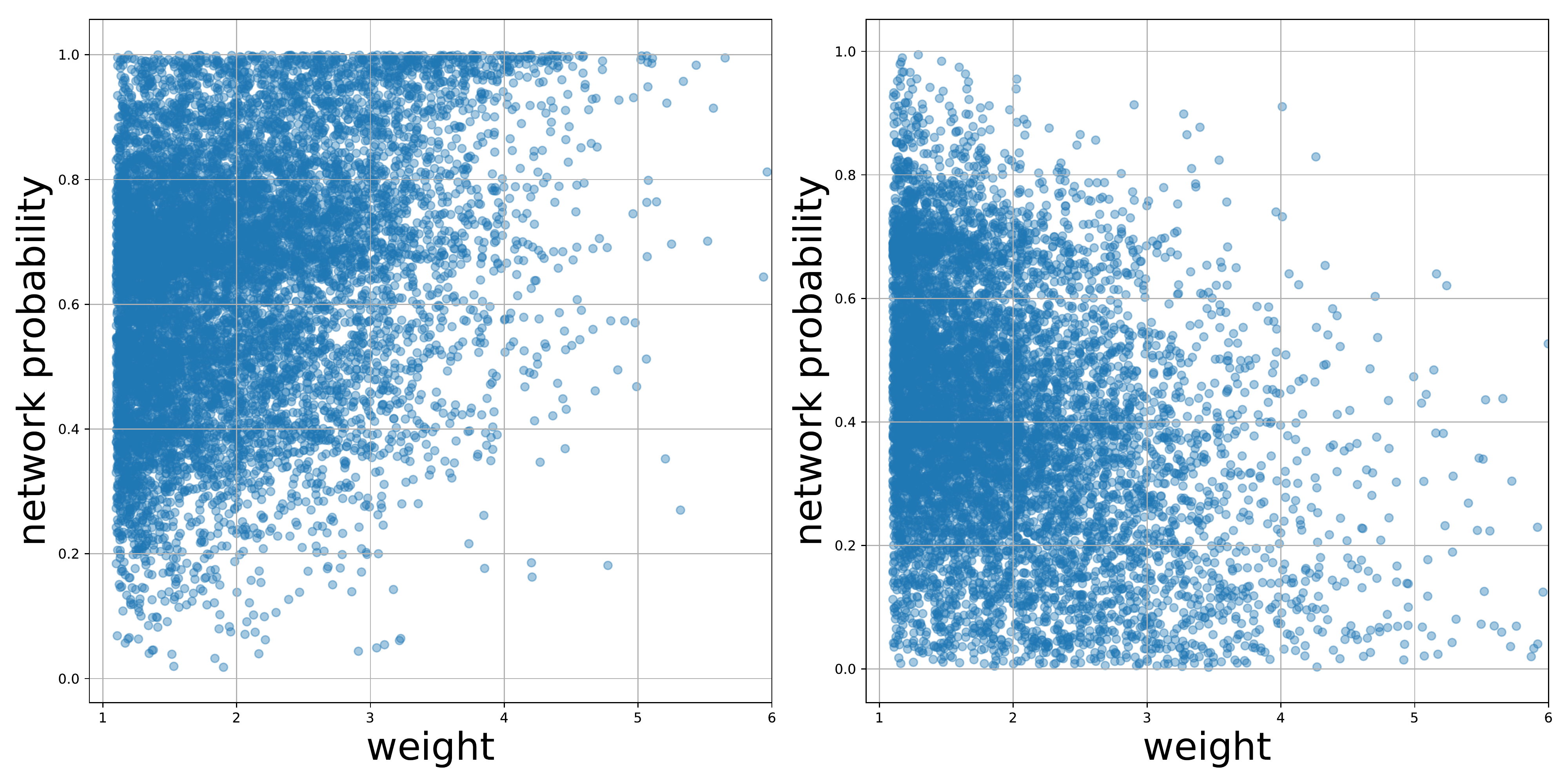}
\caption{SizeNet output probability vs. statistical weights $w_i$: Y axis is the output prediction of SizeNet. X axis is the weights $w_i$ from the statistical model corresponding to a monotonous transformation of  class confidences. Left plot is for class 1 (sizing issue) and right plot is for class 0 (no sizing issues)}
\label{fig:dots}
\end{figure}

As mentioned before, one of the added values of SizeNet is its capability to tackle the cold start problem using only images, while ensuring good performances for cases where return data is enough to accurately predict size related issues. To get a better understanding of the relation between the sample weights and the outcome of the neural network, we plot the output of the network as a function of the weights $w_i$ (a monotonous transformation of the confidence score) in \autoref{fig:dots}. In both plots, dots are distributed like triangles. Let us focus on the four corner regions of the left plot (class sizing issue) in~\autoref{fig:dots}:
\begin{itemize}
\setlength\itemsep{1mm}
    \item Upper right corner: the network outputs a value close to 1 (sizing problem) and the statistical weight is high, meaning that the teacher is very certain of the sizing issue. The dots in that area confirm that the student has learned accurately from the teacher.
    
    \item Upper left corner: the network outputs a value close to 1 (sizing problem) but the weight is low, meaning that the teacher is unsure of the class. This is the interesting case where the student correctly predicts the class, thanks to the learned visual cues, whereas the teacher fails due to lack of historic data - this region mainly corresponds to the cold start problem.
    
    \item Lower left corner: the network misclassifies samples for which the teacher is not certain of the class. Though we would prefer avoiding misclassification, those samples are next to the decision boundary where we expect disagreements between the teacher and the student.
    
    \item Lower right corner: the network misclassifies samples for which the teacher is very certain of the class. No points are observed in this region that would indicate a strong disagreement between the teacher and the student. 
        
\end{itemize}

Similar observations can be made from the right plot in~\autoref{fig:dots}, corresponding to the class 0 (no sizing issues). Following this analysis we observe that SizeNet is capable of learning and replicating the knowledge of the teacher without direct access to the privileged data. In cold-start cases, the learned cues can even help the student to make a more informed decision compared to that of the teacher. 

\subsection{Visualization of Size Issue Cues}
\label{subsec:exAI}
In the spirit of explainable AI, and to better understand the SizeNet predictions from fashion images, in this section we follow the recent methodology proposed by~\cite{petsiuk2018rise} called randomized input sampling for explanation of black-box models (RISE). We randomly generate masked versions of the input image and obtaining the corresponding outputs of the SizeNet model to assess the saliency of different pixels to the prediction. Therefore, estimated importance maps of image regions can be generated for size issue predictions in different garment categories.
\begin{figure}[ht]
   \centering
    \includegraphics[width=0.9\linewidth]{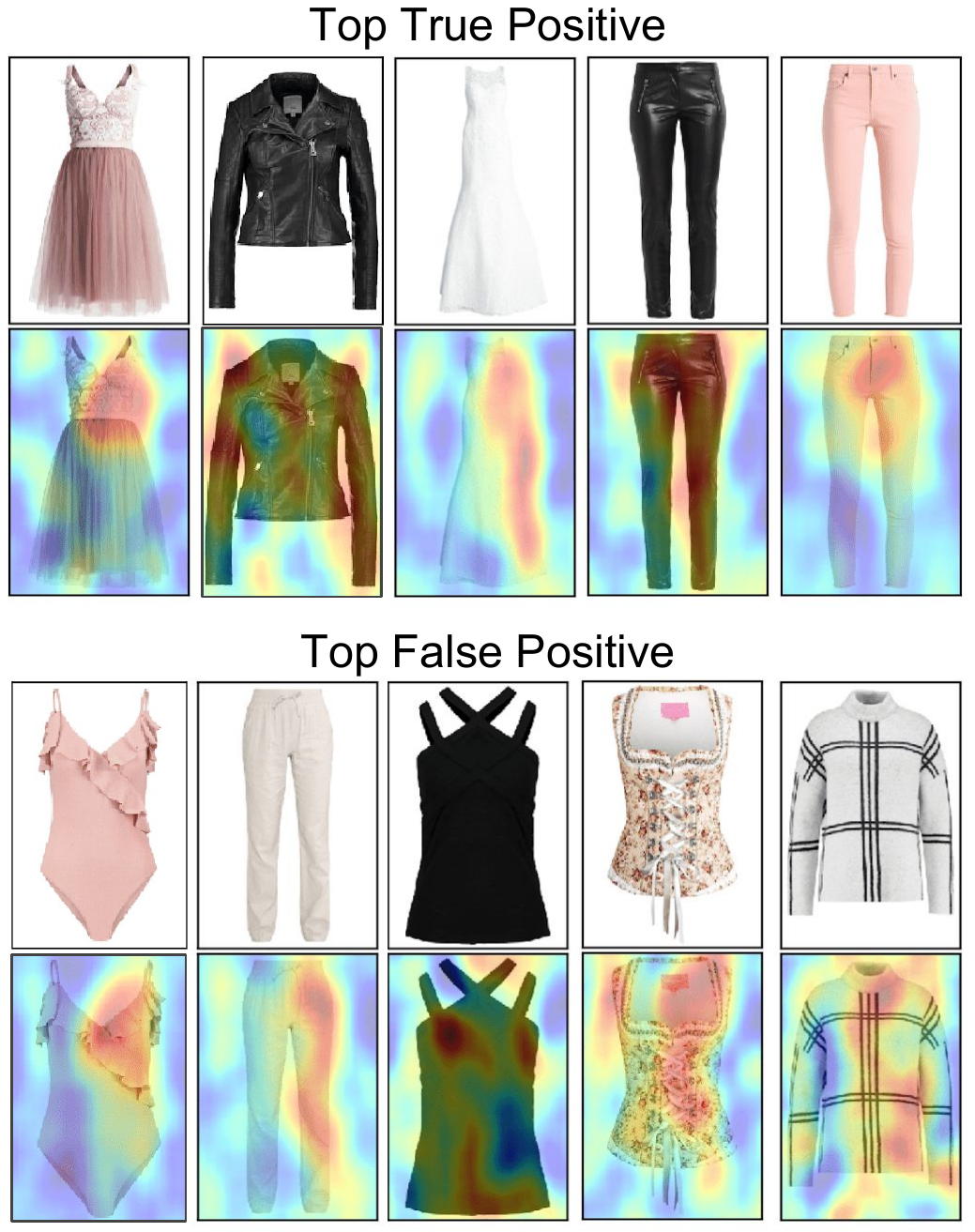}
\caption{Explanations for SizeNet model predictions: importance maps showing the effect of different image regions on the model predictions for the top five true positives (top rows), and the top five false positive predictions (bottom rows).}
\label{fig:exp_ai}
\end{figure}

In~\autoref{fig:exp_ai} we show the highest ranked true positives (top) and false positives (bottom) for size issues from different categories. It should be recalled, SizeNet was trained without any size and fit related image segment-annotations or human expert-labels. Overall from~\autoref{fig:exp_ai} we observe, for true positives, localized heatmaps attached to specific regions of the cloths, whereas for false positives we observe more expanded heatmaps covering large portions of the images. When looking closer, we can speculate that SizeNet predicts the following size issues for the highest ranked true positive articles; chest area for the evening dress, sleeves for the leather jacket, the length of the wedding dress, and areas of the trousers that may indicate too tight fit. In future work we aim to validate or reject these observations either by analyzing customer reviews on the same articles, or by including region based expert size issue annotations. On the other hand, when considering the top ranked false positives, we can observe that SizeNet misclassifies the pink top and the loose trousers based on regions of the article that are not related to size issues. These false positive examples can provide qualitative insights into the complexity of size and fit in fashion and show limitations of our approach which in its current implementation does not take into account any information on the style of fashion articles.

%% file: vSize/sections/conclusion.tex
The potential of fashion images for discovering size and fit issues was investigated. A weakly-supervised teacher-student approach was introduced where a CNN-based architecture called SizeNet, acts as a student, learns visual sizing cues from fashion images thanks to a statistical model, acting as a teacher, with privileged access to articles sales and returns data. Quantitative and qualitative evaluation was performed over different categories of garments including dresses, knitwear, tops/blouses, and trousers for both warm and cold-start scenarios. It was demonstrated that fashion images in fact contain information about article size and fit issues and can be considered valuable assets in tackling the challenging cold start problem. Future work consists of including expert-labeled data, evaluating the generalization capacities of SizeNet to fashion images in the wild, and multi-task learning for SizeNet using fit style taxonomies.
Also, further evaluation of size issue explanations derived from SizeNet is necessary to understand, on one hand, to what extend these weakly-learned localized explanations (i.e. tight shoulders, long sleeves) correspond to the actual customer experience, and on the other hand, how these explanations may be used in the future to visually support retail customers in their purchase decision making. In a longer term effort, large-scale size and fit quality metrics can be calculated for brands using SizeNet, potentially already at the prototyping stage before mass production, which can in turn result in improved products and customer satisfaction. In the future, we aim to work towards bringing a subset of our dataset to the public domain enabling further fruitful research on the challenging topic of size and fit in fashion. 

%% file: egpaper_final.bbl
\begin{thebibliography}{10}

\bibitem{Pisut2017}
Gina Pisut and Lenda~Jo Connell.
\newblock Fit preferences of female consumers in the usa.
\newblock {\em Journal of Fashion Marketing and Management: An International
  Journal}, 11(3):366--379, 2007.

\bibitem{ujevic2005}
Darko Ujevi{\'c}, Lajos Szirovicza, and Isak Karabegovi{\'c}.
\newblock Anthropometry and the comparison of garment size systems in some
  european countries.
\newblock {\em Collegium antropologicum}, 29(1):71--78, 2005.

\bibitem{shin2007}
Su-Jeong~Hwang Shin and Cynthia~L Istook.
\newblock The importance of understanding the shape of diverse ethnic female
  consumers for developing jeans sizing systems.
\newblock {\em International Journal of Consumer Studies}, 31(2):135--143,
  2007.

\bibitem{faust2014}
Marie-Eve Faust and Serge Carrier.
\newblock {\em Designing Apparel for Consumers: The Impact of Body Shape and
  Size}.
\newblock Woodhead Publishing, 2014.

\bibitem{hu2015}
Yang Hu, Xi~Yi, and Larry~S Davis.
\newblock Collaborative fashion recommendation: A functional tensor
  factorization approach.
\newblock In {\em Proceedings of the 23rd ACM international conference on
  Multimedia}, pages 129--138. ACM, 2015.

\bibitem{arora2016}
Sagar Arora and Deepak Warrier.
\newblock Decoding fashion contexts using word embeddings.
\newblock In {\em Workshop on Machine learning meets fashion, KDD}, 2016.

\bibitem{bracher2016fashion}
Christian Bracher, Sebastian Heinz, and Roland Vollgraf.
\newblock Fashion dna: Merging content and sales data for recommendation and
  article mapping.
\newblock In {\em Workshop Machine learning meets fashion, KDD}, 2016.

\bibitem{fashion_classification}
Beatriz~Quintino Ferreira, Lu{\'i}s Ba{\'i}a, Jo{\~a}o Faria, and
  Ricardo~Gamelas Sousa.
\newblock A unified model with structured output for fashion images
  classification.
\newblock In {\em Workshop on Machine learning meets fashion, KDD}, 2018.

\bibitem{DeepFashion}
Ziwei Liu, Ping Luoa, Shi Qiu, Xiaogang Wang, and Xiaoou Tang.
\newblock Deepfashion: Powering robust clothes recognition and retrieval with
  rich annotations.
\newblock In {\em Conference on Computer Vision and Pattern Recognition
  (CVPR)}, 2016.

\bibitem{amazon}
Patricia Gutierrez, Pierre-Antoine Sondag, Petar Butkovic, Mauro Lacy, Jordi
  Berges, Felipe Bertrand, , and Arne Knudsong.
\newblock Deep learning for automated tagging of fashion images.
\newblock In {\em Computer Vision for Fashion, Art and Design Workshop in
  European Conference on Computer Vision (ECCV)}, 2018.

\bibitem{ebay}
Wei Di, Catherine Wah, Anurag Bhardwaj, Robinson Piramuthu, and Neel
  Sundaresan.
\newblock Style finder: Fine-grained clothing style detection and retrieval.
\newblock In {\em Workshop in Conference on Computer Vision and Pattern
  Recognition (CVPR)}, 2013.

\bibitem{outfit}
Takuma Nakamura and Ryosuke Goto.
\newblock Outfit generation and style extraction via bidirectional lstm and
  autoencoder.
\newblock In {\em Workshop Machine learning meets fashion, KDD}, 2018.

\bibitem{tamara}
M.~Hadi Kiapour, Xufeng Han, Svetlana Lazebnik, Alexander~C. Berg, and
  Tamara~L. Berg.
\newblock Where to buy it: Matching street clothing photos to online shops.
\newblock In {\em International Conference on Computer Vision (ICCV)}, 2015.

\bibitem{lasserre2018studio2shop}
Julia Lasserre, Katharina Rasch, and Roland Vollgraf.
\newblock Studio2shop: from studio photo shoots to fashion articles.
\newblock {\em arXiv preprint arXiv:1807.00556}, 2018.

\bibitem{abdulla2017}
G~Mohammed Abdulla and Sumit Borar.
\newblock Size recommendation system for fashion e-commerce.
\newblock In {\em Workshop on Machine Learning Meets Fashion, KDD}, 2017.

\bibitem{guigoures2018hierarchical}
Romain Guigour{\`e}s, Yuen~King Ho, Evgenii Koriagin, Abdul-Saboor Sheikh, Urs
  Bergmann, and Reza Shirvany.
\newblock A hierarchical bayesian model for size recommendation in fashion.
\newblock In {\em Proceedings of the 12th ACM Conference on Recommender
  Systems}, pages 392--396. ACM, 2018.

\bibitem{sembium2017}
Vivek Sembium, Rajeev Rastogi, Atul Saroop, and Srujana Merugu.
\newblock Recommending product sizes to customers.
\newblock In {\em Proceedings of the Eleventh ACM Conference on Recommender
  Systems}, pages 243--250. ACM, 2017.

\bibitem{Sembium2018}
Vivek Sembium, Rajeev Rastogi, Lavanya Tekumalla, and Atul Saroop.
\newblock Bayesian models for product size recommendations.
\newblock In {\em Proceedings of the 2018 World Wide Web Conference}, WWW '18,
  pages 679--687, 2018.

\bibitem{zhou2017brief}
Zhi-Hua Zhou.
\newblock A brief introduction to weakly supervised learning.
\newblock {\em National Science Review}, 5(1):44--53, 2017.

\bibitem{vapnik2013nature}
Vladimir Vapnik.
\newblock {\em The nature of statistical learning theory}.
\newblock Springer science \& business media, 2013.

\bibitem{vapnik15learning}
Vladimir Vapnik and Rauf Izmailov.
\newblock Learning using privileged information: Similarity control and
  knowledge transfer.
\newblock {\em Journal of Machine Learning Research}, 16:2023--2049, 2015.

\bibitem{wong2016sequence}
Jeremy~HM Wong and Mark~John Gales.
\newblock Sequence student-teacher training of deep neural networks.
\newblock 2016.

\bibitem{bengio2009curriculum}
Yoshua Bengio, J{\'e}r{\^o}me Louradour, Ronan Collobert, and Jason Weston.
\newblock Curriculum learning.
\newblock In {\em Proceedings of the 26th annual international conference on
  machine learning}, pages 41--48. ACM, 2009.

\bibitem{word2vec}
Tomas Mikolov, Kai Chen, Greg Corrado, and Jeffrey Dean.
\newblock Efficient estimation of word representations in vector space.
\newblock {\em Workshop in International Conference on Learning Representations
  (ICLR)}, 2018.

\bibitem{lopez2015unifying}
David Lopez-Paz, L{\'e}on Bottou, Bernhard Sch{\"o}lkopf, and Vladimir Vapnik.
\newblock Unifying distillation and privileged information.
\newblock {\em arXiv preprint arXiv:1511.03643}, 2015.

\bibitem{kullback1951information}
Solomon Kullback and Richard~A Leibler.
\newblock On information and sufficiency.
\newblock {\em The annals of mathematical statistics}, 22(1):79--86, 1951.

\bibitem{resnet}
Kaiming He, Xiangyu Zhang, Shaoqing Ren, and Jian Sun.
\newblock Deep residual learning for image recognition.
\newblock {\em 2016 IEEE Conference on Computer Vision and Pattern Recognition
  (CVPR)}, pages 770--778, 2016.

\bibitem{vu2012multilingual}
Ngoc~Thang Vu, Florian Metze, and Tanja Schultz.
\newblock Multilingual bottle-neck features and its application for
  under-resourced languages.
\newblock In {\em Spoken Language Technologies for Under-Resourced Languages},
  2012.

\bibitem{petsiuk2018rise}
Vitali Petsiuk, Abir Das, and Kate Saenko.
\newblock Rise: Randomized input sampling for explanation of black-box models.
\newblock {\em arXiv preprint arXiv:1806.07421}, 2018.

\end{thebibliography}
